%% file: main.tex
\documentclass[sigconf,prologue,table]{acmart}

\AtBeginDocument{%
  \providecommand\BibTeX{{%
    \normalfont B\kern-0.5em{\scshape i\kern-0.25em b}\kern-0.8em\TeX}}}
\usepackage[linesnumbered,ruled,vlined]{algorithm2e}
\usepackage{xcolor}

\SetCommentSty{mycommfont}

\SetKwInput{KwInput}{Input}                
\SetKwInput{KwOutput}{Output}              
\usepackage{libertine}
\usepackage{booktabs} 
\usepackage{comment}
\usepackage{subcaption}
\usepackage{adjustbox}
\usepackage{makecell}
\usepackage[flushleft]{threeparttable}
\usepackage{titlesec}
\usepackage{threeparttable}
\usepackage{array}
\usepackage{enumitem}
\newcolumntype{P}[1]{>{\centering\arraybackslash}p{#1}}
\newcolumntype{M}[1]{>{\centering\arraybackslash}m{#1}}
\newlist{myenumi}{description}{10}
\setlist[myenumi]{labelindent=\parindent, leftmargin=*, align=left}
\setlist[myenumi]{leftmargin=0pt}

\renewcommand\footnotetextcopyrightpermission[1]{} 

\setcopyright{acmcopyright}
\copyrightyear{2022}
\acmYear{2022}
\acmDOI{xx.xxxx/xxxxxxx.xxxxxxx}

\acmConference[DAC'22]{Design Automation Conference}{July 2022}{San Francisco, CA}

\newcommand{\UL}[1]{\underline{#1}}

\newenvironment{tightlist}[4][\textbullet]
{%
  \begin{list}{#1}
  {%
    \setlength{\leftmargin}{#2}
    \setlength{\rightmargin}{#3}
    \setlength{\topsep}{0pt}
    \setlength{\parsep}{0pt}
    \setlength{\listparindent}{0pt}
    \setlength{\itemsep}{#4}
  }
}
{%
  \end{list}
}

\titlespacing\section{1pt}{1pt plus 1pt minus 1pt}{0pt plus 1pt minus 1pt}
\titlespacing\subsection{1pt}{1pt plus 2pt minus 1pt}{1pt plus 2pt minus 1pt}
\titlespacing\paragraph{10pt}{1pt plus 2pt minus 1pt}{1pt plus 2pt minus 1pt}[]

\begin{document}

\title{GATSPI: GPU Accelerated Gate-Level Simulation for Power Improvement}


\author{ Yanqing Zhang,
         Haoxing Ren,
         Akshay Sridharan,
         Brucek Khailany
         }
\affiliation{
    \institution{NVIDIA Corporation}
    \institution{\{yanqingz, haoxingr, asridharan, bkhailany\}@nvidia.com}
    \country{}
}

\input{0.abstract}
\settopmatter{printfolios=true}
\maketitle

\input{1.intro}

\input{2.related_work}

\input{3.GATSPI_implementation}

\input{4.Experiments}
\input{5.Analysis}
\input{6.Conclusions}

\bibliographystyle{unsrt_abbrv_custom}
\bibliography{GATSPI_bib}


\end{document}

%% file: 0.abstract.tex
\begin{abstract}
\label{0.abstract}
In this paper, we present GATSPI, a novel GPU accelerated logic gate simulator that enables ultra-fast power estimation for industry-sized ASIC designs with millions of gates. GATSPI is written in PyTorch with custom CUDA kernels for ease of coding and maintainability.  It achieves simulation kernel speedup of up to 1668X on a single-GPU system and up to 7412X on a multiple-GPU system  when compared to a commercial gate-level
simulator running on a single CPU core. GATSPI supports a range of simple to complex cell types from an industry standard cell library and SDF conditional delay statements without requiring prior calibration runs and produces industry-standard SAIF files from delay-aware gate-level simulation. Finally, we deploy GATSPI in a glitch-optimization flow, achieving a 1.4\% power saving with a 449X speedup in turnaround time compared to a similar flow using a commercial simulator.

\end{abstract}

%% file: 1.intro.tex
\section{Introduction}
\label{1.intro}
\input{figs/Figs_ProblemDescription.tex}
Gate-level logic simulation plays an important role in the design and signoff of integrated circuits. Simulation is used in many steps such as power analysis, design-for-test (DFT) pattern generation, DFT power analysis, and fault simulation. However, gate-level simulation, and especially delay-accurate glitch-enabled simulation for power estimation, can exhibit long runtimes on conventional simulators, prohibiting its widespread adoption in chip design flows. Today's highly complex SoCs have hundreds of partitions, each with millions of gates and numerous testbenches running for thousands of cycles to simulate.  Accurate results can take many hours to days across hundreds of servers to produce. Thus, there is a desire to accelerate this simulation task. Faster alternatives such as probability-based switching activity estimators can also produce toggle counts/rates for power purposes, but they are inaccurate. 

Typically, when running delay-aware and glitch-enabled gate-level simulation, correct waveforms of sequential elements are already known in advance from previous RTL simulation results, automatic test pattern generation (ATPG) vectors, or scan test. These cases, where the input stimuli to the logic cones in the digital design from primary inputs and pseudo-primary inputs (register and RAM outputs) are known, have been called logic `re'-simulation\cite{ProblemC} (Fig.~\ref{Figs_ProblemDescription}(Left)). Logic re-simulation is beneficial for turnaround time, since a re-simulator can forego functional simulation of the sequential elements. What's more, in re-simulation, the known sequential element waveforms break the cycle-to-cycle dependency so that multiple cycles of the testbench can be simulated in parallel. Modern day parallel computing architectures, such as GPUs, provide an opportunity for speedups\cite{Holst2015,Zhu2011,Sen2010,Chatterjee2011,Lai2020,Zeng2021} through parallelization across cycles (stimuli) as well as the design (gates within the same logic level can be simulated in parallel), as depicted in  Fig.~\ref{Figs_ProblemDescription}(Right). 

\input{figs/Table_GPUarchs.tex}
However, GPUs also present challenges to re-simulation acceleration, such as the development cost for writing and maintaining a separate code base in GPU programming languages such as CUDA. Other challenges include minimizing the communication overhead between host and device (CPU and GPU), efficient scheduling, and avoiding memory bandwidth bottlenecks to keep the Streaming Multiprocessors (SMs) occupied with sufficient data. Fortunately, newer GPUs have features that help overcome these challenges, most notably plentiful memory bandwidth and expanded GPU memory capacity that rivals some system memory capacities (Table~\ref{Table_GPUarchs}). 

In addition, the emergence of deep learning (DL) has helped foster a rich software ecosystem for programming GPUs. Popular DL frameworks such as PyTorch\cite{PyTorch} lower the barrier for developing GPU code that operates efficiently on matrices and tensors. Graph learning packages such as DGL\cite{DGL}, built on top of PyTorch, provide APIs for various graph algorithms and partitioning methods. Since digital logic circuits can be naturally represented as graphs, these frameworks can be exploited for productive GPU programming on logic re-simulation and other related tasks.

In this work, we build upon these recent developments in GPU software and hardware platforms
with a new GPU-accelerated gate-level re-simulator called GATSPI (which stands for \UL{G}PU \UL{A}ccelerated Ga\UL{T}e-level \UL{S}imulation for \UL{P}ower \UL{I}mprovement). The novel contributions of this work are:
\begin{tightlist}{1em}{1em}{0em}
    \item[$\bullet$] GATSPI achieves up to 1668X re-simulation kernel speedup on a single A100 GPU and up to 7412X on 8 V100 GPUs when compared to a commercial  simulator across a diverse set of industry benchmarks with no accuracy loss. It also achieves up to 680X speedup when considering end-to-end application tasks, such as waveform loading and writing results to file.
    \item[$\bullet$] To the best of our knowledge, GATSPI is the first GPU accelerated re-simulator that supports full logic cell types, all SDF conditional delay statements, multiple-simultaneous-input (MSI) switching resolution, and inertial delay filtering on both gates and interconnect--with no need for calibration runs.
    \item[$\bullet$] For ease of coding, GATSPI is novelly implemented in PyTorch, DGL, and with only 2 PyTorch custom CUDA kernels.
    \item[$\bullet$] Through GPU profiling, we show analysis for `hyperparameter' tuning (amount of cycle-level parallelism, registers/thread, threads/block) for the GATSPI kernels. 
    \item[$\bullet$] We deploy GATSPI in a glitch-optimization flow, achieving a 1.4\% power saving with a 449X speedup in turnaround time compared to a similar flow using a commercial CPU-based simulator.
\end{tightlist}



%% file: figs/Figs_ProblemDescription.tex
\begin{figure}[t]
  \setlength{\abovecaptionskip}{3pt}
  \setlength{\belowcaptionskip}{3pt}
  \centering
  \includegraphics[width=1\columnwidth]{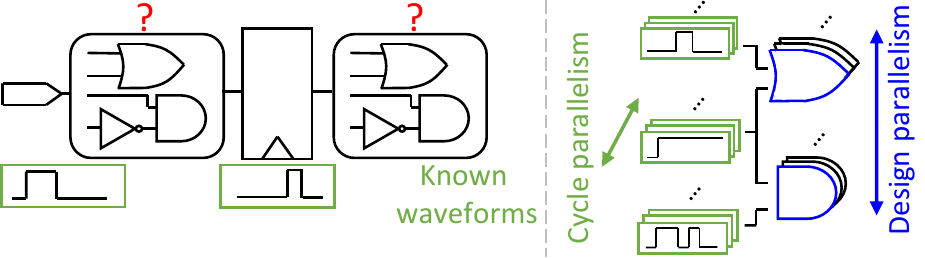}
  \caption{\normalfont{(Left) Depiction of re-simulation. Only logic gates need to be simulated. (Right) Parallelism opportunities in re-simulation, which can be mapped to GPUs.}
  \label{Figs_ProblemDescription}}
\end{figure}


%% file: figs/Table_GPUarchs.tex
\begin{table}[t]
  \centering
  \captionsetup{belowskip=0pt,aboveskip=3pt}
  \caption{\normalfont{Comparison of recent NVIDIA GPU architectures\cite{GPUarchs,T4arch}.}}
    \resizebox{\columnwidth}{!}{\begin{tabular}{cccc}
    \textbf{Architecture} & \textbf{T4} & \textbf{V100} & \textbf{A100} \\
    \midrule
    \textbf{SMs} & 40 & 80 & 108 \\
    \textbf{Global Memory Size} & 16 GB & 16 or 32 GB & 40 or 80 GB \\
    \textbf{Memory BW} & 320 GB/s & 900 GB/s & 1.6 or 1.9 TB/s \\
    \textbf{L2 cache} & 4 MB  & 6 MB  & 40 MB \\
    \bottomrule
    \end{tabular}}
  \label{Table_GPUarchs}%
    \vspace{0pt}
\end{table}%


%% file: 2.related_work.tex
\section{Related Work}
\label{2.related_work}
Prior work has explored GPU accelerated gate-level simulation. Early efforts explored a mostly \textit{oblivious simulator} approach, evaluating every gate at every cycle or timestamp regardless of input activity \cite{Sen2010,Chatterjee2011}. These oblivious simulators are computationally inefficient at low activity factors, and require massive amounts of memory to log every gate at every timestamp, achieving 2-45X \textit{simulation kernel} (only actual simulation runtime is measured) speedup \cite{Sen2010,Chatterjee2011}. Another approach used an \textit{event-based simulator} approach and a sophisticated memory paging technique to perform re-simulation for 1-270X speedup \cite{Zhu2011}. Event-based simulation only performs evaluation on specific gates when changes on their inputs occur. As stated in \cite{Zhu2011}, the speedups depend on the activity factor of the testbench.

The best `class' of GPU accelerated re-simulators have been \textit{hybrid event/oblivious simulators}\cite{Holst2015, Zeng2021, Lai2020}. The general idea of hybrid approaches is that each gate's simulation advances in time based on events on its inputs, but the scheduling and launching of threads on the GPU for each gate is oblivious. Each gate will always have a simulation thread assigned to it, and threads whose gate has little or no activity simply finish earlier. One recent work implemented such an approach on industry-sized designs with multi-millions of gates and achieved 23-44X speedup on end-to-end \textit{application} runtimes (measured from loading the testbench waveforms until result file dumping) \cite{Lai2020}. Holst et al. implemented a hybrid 2-value re-simulator with a novel memory management scheme which requires calibration loops, and they achieved large simulation kernel speedups of 21-1090X for long-duration scan testbenches on experimental(<550k 2-input only gates) circuits \cite{Holst2015}.  Another recent work implemented a 4-value re-simulator that achieved 2-44X application speedup on the ICCAD 2020 Design Contest\cite{ProblemC} benchmarks~\cite{Zeng2021}. The contest\cite{ProblemC} asks for an implementation of transport delays instead of inertial delays, and \cite{Zeng2021} cleverly used this deviation in gate delay processing to implement another degree of parallelism -- event parallelism. They simulate each intra-cycle event in parallel. This is unique to transport delay processing because there is no pulse filtering, so same-value output events and delay collision events can be invalidated after the parallel simulation of individual events is completed.

These hybrid simulators typically partition the design netlist by logic level, which is a convenient and straightforward way to access design parallelism \cite{Holst2015,Zeng2021}, and GATSPI similarly adopts this approach. Simulation only advances to the next logic level partition if the current level has completed, which ensures input/output waveform dependencies are always met. In comparison to \cite{Holst2015,Zhu2011}, GATSPI aims to use the expanded memory capabilities of newer GPUs without memory management schemes. Similar to \cite{Holst2015}, GATSPI implements 2-value re-simulation, since gate re-simulation for power analysis scarcely produce x/z values.  However, \cite{Holst2015} only supports 2-input gates, whereas GATSPI supports all boolean logic gate types and conditional SDF delay statements. These features are important for targeting industry-standard libraries. In contrast to \cite{Zeng2021}, GATSPI processes inertial delays. While simpler to implement, transport delay processing will lead to overestimates of glitches, the inaccuracy of which is not conducive in power analysis settings.

Finally, similar to DREAMPlace \cite{Lin2019}, GATSPI leverages DL framework software packages such as PyTorch and DGL along with several custom CUDA kernels to ease the software development and maintenance effort.

%% file: 3.GATSPI_implementation.tex
\section{GATSPI Implementation}
\label{3.GATSPI_implementation}
\input{figs/Figs_ToolFlow.tex}
\input{figs/Figs_Waveform.tex}
Fig.~\ref{Figs_ToolFlow} shows the overall application workflow of GATSPI compared to a commercial simulator flow. Our starting points are a gate-level netlist, corresponding SDF file, and primary/pseudo-primary input waveforms. 

We employ the same waveform format as \cite{Holst2015} for efficient storage, shown in Fig.~\ref{Figs_Waveform}. We use a Python script to translate SDF delay statements and logic function truth tables into the array format described in Fig.~\ref{Figs_WaveformLUTSDF}. We translate the netlist into a PyTorch/DGL graph object in a manner similar to \cite{GRANNITE}. The DGL graph object retains netlist information such as connectivity, gate and interconnect delays (edge features), and cell logic function (node feature) by using DGL's graph node/edge attribute annotation. Finally, GATSPI loads the DGL graph object and input array waveforms, performs logic levelization, and launches the CUDA kernels to complete simulation. The resulting SAIF file for downstream power analysis or other applications can be dumped asynchronously as the CUDA kernels are running \cite{Zeng2021}.
\input{figs/Figs_WaveformLUTSDF.tex}
\input{figs/Figs_Simulator.tex}

Fig.~\ref{Figs_Simulator} provides an overview of GATSPI's simulation approach. It first loads the input waveforms and pre-allocates one chunk of device memory for storing \textbf{all} the waveforms of the simulation. For example, when using an NVIDIA V100 GPU with 32GB device memory, we allocate 24 GB of memory to store waveforms--the remaining 8GB is used to store the DGL graph and input/output waveform address pointers. This approach avoids all data movement between CPU host and GPU device (aside from asynchronous dumping of results to file) during simulation, removing communication bottlenecks. For each logic level, each gate/cycle to be simulated is assigned to a thread, and each thread `fetches' its respective input waveforms from device memory by referencing the corresponding waveform's start-address pointer. The thread hierarchy for each simulation kernel is designed to match the cycle+design parallelism schema in Fig.~\ref{Figs_ProblemDescription}. However, this introduces a gate output waveform addressing issue: we don't know a priori what the output waveform's start-address pointer value should be prior to simulation, and using dynamic arrays would be slow. One solution is to use calibration loops\cite{Holst2015}, but this may increase latency with repeated simulation. Instead, GATSPI simply simulates twice\cite{ICCAD2020Winner}: once to determine the output toggle count (TC) which determines the address value of the output waveform pointers, and again to perform the same re-simulation whilst storing the waveform.

GATSPI is conveniently written in PyTorch and DGL, which has many advantages. Many issues such as indexing of multi-dimension arrays are made convenient through PyTorch, and graph netlist manipulation is made convenient through DGL. Only the re-simulation kernels of Algo. 1 are written in CUDA, integrated as custom PyTorch CUDA calls. Each thread in the kernel simulates one independent cycle/stimuli of one gate in the current logic level. Algo. 1 implements a delay-aware algorithm for logic gate simulation with inertial delay filtering on both gates and wire interconnect, processing of conditional SDF delay statements, and MSI switching resolution.
\input{figs/AlgoTest.tex}

Algo. 1 describes the second iteration of re-simulation in GATSPI where the TC values are known and the output waveforms are to be stored in memory. The first iteration's algorithm is very similar. After resolving the gate's initial value, lines 8-13 determine the timestamp of the next earliest input transition to process. Of note, lines 10-12 perform interconnect inertial delay filtering. Also of note, lines 8-13 only determine the next timestamp, \textbf{not} which pin is switching. This is because lines 14-18 loop through all input pins to find all MSI switching conditions, and resolve them before updating gate output value \textit{y} and \textit{gateDelay}. Algo. 1 will only trigger its inertial delay filtering logic (lines 21-25) if a change on output value \textit{y} is detected (line 19). Algo. 1 assumes a simulation constraint of PATHPULSEPERCENT=100. Other constraint values can be easily implemented by editing line 21.

We followed the CUDA tuning principles outlined in~\cite{CUDAtuning} where possible during implementation of GATSPI's CUDA kernels. As a result, we make use of asynchronous memory copy and we pin the pointer variables $p_o$ and $p_i$ to registers for 2\% kernel speedup compared to kernel performance prior to tuning.


%% file: figs/Figs_ToolFlow.tex
\begin{figure}[t]
  \setlength{\abovecaptionskip}{0pt}
  \setlength{\belowcaptionskip}{0pt}
  \centering
  \includegraphics[width=1\columnwidth]{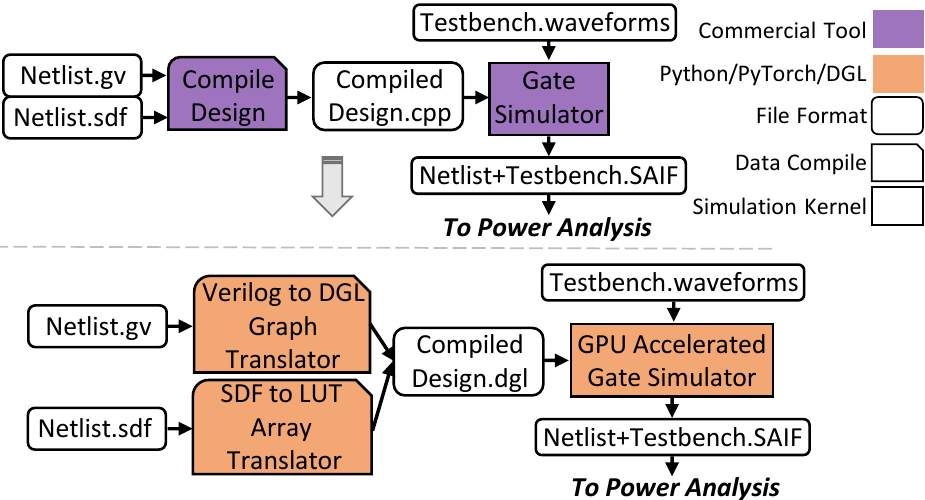}
  \caption{\normalfont{Overview of tool flow for commercial tools and GATSPI.}
  \label{Figs_ToolFlow}}
\end{figure}


%% file: figs/Figs_Waveform.tex
\begin{figure}[t]
  \setlength{\abovecaptionskip}{0pt}
  \setlength{\belowcaptionskip}{0pt}
  \centering
  \includegraphics[width=1\columnwidth]{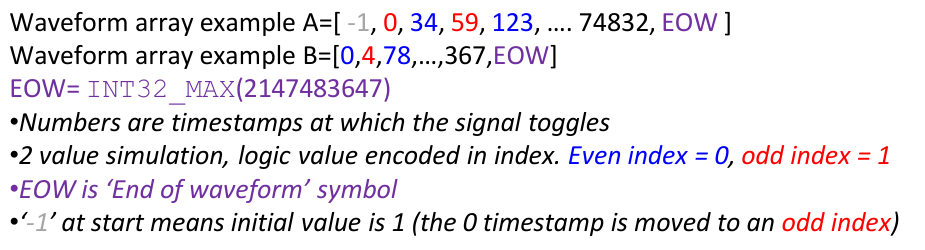}
  \caption{\normalfont{Array format of signals in GATSPI taken from \cite{Holst2015}.}
  \label{Figs_Waveform}}
\end{figure}


%% file: figs/Figs_WaveformLUTSDF.tex
\begin{figure}[htbp]
  \setlength{\abovecaptionskip}{0pt}
  \setlength{\belowcaptionskip}{4pt}
  \centering
  \includegraphics[width=1\columnwidth]{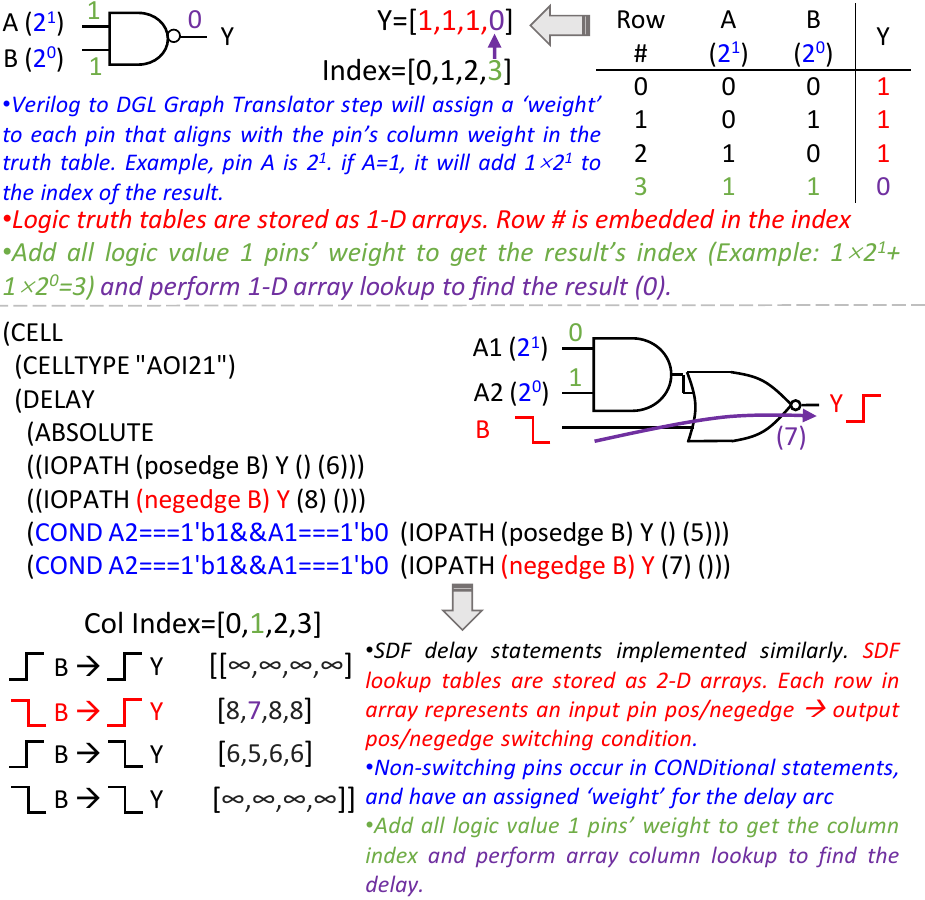}
  \caption{\normalfont{GATSPI formulates logic function evaluation and arc delay determination as uniform array lookup operations to support a variety of logic cell types and conditional delay statements.}
  \label{Figs_WaveformLUTSDF}}
\end{figure}
\vspace{0pt}

%% file: figs/Figs_Simulator.tex
\begin{figure}[htbp]
    \setlength{\abovecaptionskip}{0pt}
  \setlength{\belowcaptionskip}{2pt}
  \centering
  \includegraphics[width=1\columnwidth]{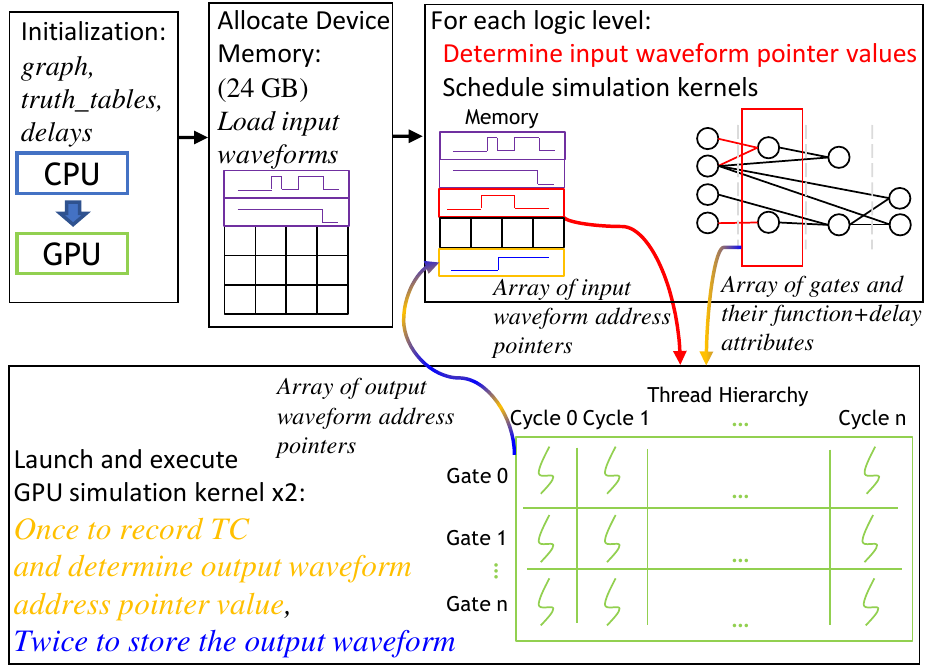}
  \caption{\normalfont{Simulation Overview.}
  \label{Figs_Simulator}}
\end{figure}

\vspace{0pt}


%% file: figs/AlgoTest.tex
\begin{algorithm}[t]
\caption{Per-gate, per-cycle parallelism re-simulation}
\DontPrintSemicolon
  \textbf{Input}: 1-D array of all waveforms $allW$. Number of pins $n$. 1-D array of logic truth table(Fig.~\ref{Figs_WaveformLUTSDF}) $Y\in [2^{n}]$. Per input pin $i$ rise/fall wire delays $\delta_i=[\delta^{r}_i, \delta^{f}_i]$. 2-D array of gate delays(Fig.~\ref{Figs_WaveformLUTSDF}) $d_i\in [4,2^{n-1}]$. Per pin $i$ waveform address pointer $p_i$. Gate output waveform address pointer $p_o$.
  
  \textbf{Output}: Output waveforms stored in $allW$.
  
  \textbf{For each $i$}: $allW[p_i]==-1$ $?$ $p_i++$ $:$ $null$
  \tcc{initial value}
  
  $colInd = \sum_{i=1}^{n}(p_i \% 2) * 2^{i-1}$ \tcc{\% is the modulo operator}
  $y = Y[colInd]$
  
  $y==0$ $?$ $allW[p_o]=0$ $:$ ($allW[p_o]=-1$ $;$ $allW[p_o++]=0$)

  \While{$t_i != EOW$} 
   {
   		$t_i = EOW$ \tcc{EOW is INT\_MAX(Fig.~\ref{Figs_Waveform})}
   		
   		\For{\normalfont{\textbf{each}} $i$}
   		{ 
        	$netDelay = (p_i \% 2 )$ $?$ $\delta^{f}_i$ $:$ $\delta^{r}_i$ $;$ 
        	
        	\If{$allW[p_i+2] - netDelay - allW[p_i+1] < 0$}
            {
                 $p_i += 2 ; continue$
            }
            
            $t_i =$ min( $t_i$, $allW[p_i+1] + netDelay$)
        }
        \If{$t_i!=EOW$}
        {
            \For{\normalfont{\textbf{each}} $i$}
            {
                \If{\normalfont{transition time of pin} $i==t_i$}
                {
                    $p_i++$
                    
                    update $colInd;$ update $y;$ set $gateDelay$ indexing $d_i$ using $p_i\%2$ and $colInd$
                }
            }
            \If{$y!=p_o\%2$}
            {
                $t_o = t_i + gateDelay$
                
                \If{$t_o - allW[p_o] < gateDelay$}
                {
                    $p_o--$
                }
                \Else
                {
                    $p_o++$
                    
                    $allW[p_o]=t_o$
                }
            }
        }
   }
\label{Algo_Simulation}
\end{algorithm}

%% file: 4.Experiments.tex
\section{Experiments and Results}
\input{figs/Table_benchmarks.tex}
\input{figs/Table_OpenMP}
\input{figs/Table_FGP}
\label{4.Experiments}
We perform GPU accelerated gate simulation with GATSPI on the benchmarks listed in Table~\ref{Table_benchmarks}. Benchmarks were chosen to provide diversity in design netlist size/structure, testbench activity factor, and simulation length. Of note, many benchmarks are chosen from industrial-scale designs with millions of gates and testbenches from power windows that informed power signoff for tapeout.  We compare GATSPI's execution time against the baseline of a commercial simulator, which is a commonly used baseline~\cite{ProblemC, Holst2015, Lai2020, Zeng2021}. Both application runtime\cite{Lai2020,Zeng2021}(i. e. contents of Fig.~\ref{Figs_Simulator}) and simulation kernel runtime\cite{Holst2015}(i. e. contents of Algo. 1) is compared. Simulation correctness is verified in two ways: the resulting SAIF files are compared, and `spot-checks' of full waveforms of some random signals are compared for each benchmark between baseline and GATSPI. All GATSPI experiments in Table~\ref{Table_benchmarks} were performed on a system with a 32 GB NVIDIA Quadro GV100 and 30 GB system memory. Cycle-parallelism is set to 32 (32 indepedent cycles are re-simulated in parallel). Baseline experiments with a commercial simulator were performed on an Intel Xeon single E5@2.70GHz CPU core with 64 GB system memory. Re-simulation is implemented in the commercial tool by making use of the \textit{force} keyword of SystemVerilog.

Table~\ref{Table_benchmarks} also reports the activity factor of each benchmark.
Hybrid simulators on GPUs tend to have a maximum throughput measured in total \textit{events simulated/second}, so including the activity factor metric provides a clearer assessment of throughput achieved. GATSPI achieves 28-1198X simulation kernel speedup, or 5-680X application speedup across a diverse set of design size, testbench activity factor, and testbench lengths. The application overheads, such as re-structuring the input waveforms to match the cycle-parallelism schema and dumping the result file, take up most of the application runtime. Unsurprisingly, largest speedups are achieved for long durations of high activity, since more time is spent within the re-simulation kernels, and simulation speedup is not hurt as much by Amdahl's law. Though not experienced in our benchmarks, in cases where the number of re-simulation events to be stored in device memory surpass its capacity, the testbench can be compiled into shorter segments, GATSPI can be invoked several times in sequence, and dump files can be combined to achieve the full simulation results.

We also evaluate GATSPI in a multi-threaded CPU environment by modifying GATSPI's code to support an equivalent OpenMP implementation. For brevity, we choose a few typical representative benchmarks to perform these experiments on (Table~\ref{Table_OpenMP}). We consider the three listed benchmarks as representative: a small design whose workload barely saturates the max thread count on our GPU platform, an industrial design with long testbench but low activity and highly unbalanced workload between different gate instances/threads, and an industrial design with long testbench and relatively high activity and somewhat balanced workload between different gate instances/threads. Similarly, we compare GATSPI against a commercial simulator's multi-threaded CPU implementation (Table~\ref{Table_FGP}) running on Intel Xeon E5 CPUs in a server farm. In this case, benchmarks are chosen because they did not segfault. GATSPI substantially outperforms both scenarios.

Finally, to demonstrate the utility of GATSPI in a VLSI optimization flow, we incorporated GATSPI into a glitch-power-reduction tool flow on an industry design. We first re-simulated a 1.3M gate design to attain its activity. Custom scripts then perform glitch activity/design analysis before designer-informed glitch-fixing transformations are made to the netlist. Often glitch fixes at one testbench can lead to power changes in other testbenches. The updated netlist is then re-simulated to assess power savings. GATSPI reduced the turnaround time of re-simulation from 1459.6 minutes to 3.25 minutes, a 449X speedup, confirming the 1.4\% design (1\% whole chip) power savings. Such fast turnaround times can improve chip designers' productivity and enable future automation. Furthermore, GATSPI's fast simulation throughput can enable efficient mass simulation data collection for other use cases requiring accurate delay-annotated gate simulation. Examples may include crosstalk analysis, IR drop analysis, power-aware logic synthesis, or DFT.


%% file: figs/Table_benchmarks.tex
\begin{table*}[t]
\centering
\captionsetup{belowskip=0pt,aboveskip=0pt}
\caption{\normalfont{Benchmarks and results on V100. Industry designs in grey. Open source benchmarks from \cite{NVDLA}. Speedups in parentheses.}}
\label{Table_benchmarks}
\resizebox{\textwidth}{!}{
    \begin{tabular}{ccccc|cc|cc}
    \textbf{Design (Configuration)} & \textbf{Testbench} & \multicolumn{1}{M{3.445em}}{\textbf{Gate Count}} & \multicolumn{1}{M{3.445em}}{\textbf{Activity Factor}} & \multicolumn{1}{p{2.61em}|}{\textbf{Cycles}} & \multicolumn{1}{M{6.035em}}{\textbf{Baseline App. Runtime(s)}} & \multicolumn{1}{M{8.235em}|}{\textbf{Baseline Re-sim. Kernel Runtime(s)}} & \multicolumn{1}{M{5.78em}}{\textbf{GATSPI App. Runtime(s)}} & \multicolumn{1}{M{8.22em}}{\textbf{GATSPI Re-sim. Kernel Runtime(s)}} \\
    \midrule
    \textbf{32b\_int\_adder} & random stimulus & 1k    & 1.0   & 60k   & 554   & 529   & 5.98 (93X) & 5.75 (92X) \\
    \textbf{NVDLA\_m(small)} & convolution & 14k   & 0.058 & 743k  & 455   & 373   & 12.05 (38X) & 4.35 (86X) \\
    \textbf{NVDLA\_m(large)} & convolution & 257k  & 0.0017 & 132k  & 159   & 133   & 8.56 (19X) & 1.4 (95X) \\
    \textbf{NVDLA\_m(large)} & scan  & 257k  & 1.2   & 5k    & 723   & 670   & 18.27 (40X) & 3.82 (175X) \\
    \textbf{NVDLA(large)} & sanity test & 1.8M  & 0.00079 & 100k  & 180   & 116   & 35.41 (5X) & 4.09 (28X) \\
    \textbf{NVDLA(large)} & scan  & 1.8M  & 1.0   & 1.5k  & 3211  & 2535  & 70.81 (45X) & 9.99 (254X) \\
    \rowcolor[rgb]{ .682,  .667,  .667} \textbf{Industry Design A} & functional 1 & 77k   & 0.094 & 9.4k  & 670   & 635   & 4.05 (165X) & 0.79 (808X) \\
    \rowcolor[rgb]{ .682,  .667,  .667} \textbf{Industry Design B} & functional 2 & ~2M   & 0.013 & 78k   & 16060 & 14924 & 41.76 (385X) & 14.55 (1026X) \\
    \rowcolor[rgb]{ .682,  .667,  .667} \textbf{Industry Design B} & high activity short test & ~2M   & 0.186 & 11k   & 20969 & 18727 & 53.46 (392X) & 19.18 (976X) \\
    \rowcolor[rgb]{ .682,  .667,  .667} \textbf{Industry Design B} & high activity long test & ~2M   & 0.183 & 33k   & 49230 & 46617 & 72.35 (680X) & 38.90 (1198X) \\
    \rowcolor[rgb]{ .682,  .667,  .667} \textbf{Industry Design C} & functional 2 & ~1.9M & 0.015 & 32k   & 6224  & 5065  & 38.91 (160X) & 6.98 (726X) \\
    \rowcolor[rgb]{ .682,  .667,  .667} \textbf{Industry Design D} & functional 3 & ~2.3M & 0.024 & 62k   & 10638 & 8896  & 68.12 (156X) & 15.72 (566X) \\
    \bottomrule
    \end{tabular}%
}
\vspace{-15pt}
\end{table*}

%% file: figs/Table_OpenMP.tex
\begin{table}[t]
  \centering
  \captionsetup{belowskip=0pt,aboveskip=0pt}
  \caption{\normalfont{GATSPI comparison to its OpenMP implementation. Speedups vs. OpenMP in parentheses.}}
    \resizebox{\columnwidth}{!}{\begin{tabular}{cccc}
    \textbf{Design(Testbench)} & \multicolumn{1}{M{7.61em}}{\textbf{GATSPI Kernel Runtime (s)}} & \multicolumn{1}{M{7.72em}}{\textbf{OpenMP Kernel Runtime (s)}} & \multicolumn{1}{M{3.89em}}{\textbf{\# CPUs Used}} \\
    \midrule
    Design A(func. 1) & 0.79(12.8X)  & 10.10 & 32 \\
    Design B(func. 2) & 14.55(9.4X) & 136.09 & 40 \\
    Design B(high activity) & 38.90(14.4X) & 558.94 & 64 \\
    \bottomrule
    \end{tabular}}
  \label{Table_OpenMP}%
    \vspace{2pt}
\end{table}%


%% file: figs/Table_FGP.tex
\begin{table}[t]
  \centering
  \captionsetup{belowskip=0pt,aboveskip=0pt}
  \caption{\normalfont{GATSPI comparison to multi-threaded commercial tool. Speedups vs multi-threaded tool version in parentheses.}}
    \resizebox{\columnwidth}{!}{\begin{tabular}{cccc}
    \textbf{Design(Testbench)} & \multicolumn{1}{M{5.835em}}{\textbf{GATSPI App. Runtime (s)}} & \multicolumn{1}{M{6.055em}}{\textbf{Baseline App. Runtime (s)}} & \multicolumn{1}{M{6.345em}}{\textbf{Multi-thread Runtime (s)}} \\
    \midrule
    Design A(func. 1) & 4.05(63.7X)  & 670   & 258 \\
    NVDLA\_m,large(scan) & 18.27(50.0X) & 3211  & 914 \\
    \bottomrule
    \end{tabular}}
  \label{Table_FGP}%
    \vspace{0pt}
\end{table}%

%% file: 5.Analysis.tex
\section{GPU Profiling and Analysis}
\label{5.Analysis}

\textit{\textbf{GPU Profiling}}~--- We used NVIDIA's Nsight~\cite{nsys} to profile the GATSPI full application and simulation kernel to provide insight into achieved performance and identify areas of possible improvement. We again used the 3 representative benchmarks for brevity in our analysis experiments. Table~\ref{Table_NSYS} shows the results of application profiling, which reveal that data loading from host to device is not a major contributor to overall application runtime. Most of the initialization stage of GATSPI is spent re-structuring the input waveforms' format, needed for exploiting cycle parallelism. There is also a static cost for stream synchronization and kernel launch time based on the logic level structure of the netlist, determined by the logic level partitioning.

Table~\ref{Table_NCU} shows kernel profiling results for the second GATSPI CUDA kernel where output waveforms are stored in memory, and for the widest logic level in the design. Several bottlenecks can be identified.  First, Algo. 1 contains many loops and memory access is highly irregular. Though high memory bandwidth is beneficial, compute throughput is restricted by waiting for memory responses (low L2 Hit Rate, high Cycle per Issue). Memory throughput is restricted by the algorithm fetching singular data (the next transition timestamp on each non-consecutively stored input pin waveform), as evidenced by uncoalesced memory accesses.

Kernel profiling also informs GPU application `hyperparameter' tuning--in the case of GATSPI, these parameters include: amount of cycle parallelism, threads/block, and registers/thread.  For reasons we explain below, we chose a configuration of \{32,512,64\} for parameters \{Cycle parallelism, threads/block, registers/thread\} to optimize GATSPI runtime.

For cycle parallelism, we undoubtedly wish to set this parameter to at least 32. There are 32 threads to a \textit{warp} on the GPU, and instructions are issued per warp. So, setting cycle parallelism to a multiple of 32 causes each warp to simulate one gate instance, which helps limit code divergence. We may assume the greater the cycle parallelism, the faster the runtime, but Table~\ref{Table_NCU} reveals otherwise. In the case where the design is small, initially higher cycle parallelism lowers latency. But, eventually the L2 reaches capacity, and its hit rate decreases leading to longer stalls and increased latency. In the more typical case of large industrial-scale designs, higher cycle parallelism leads to denser issuing of memory instructions that memory cannot fully keep up with, which drastically lowers memory throughput, leading to increased latency. GATSPI sets cycle parallelism to 32, sacrificing some flexibility for robust behavior.

A CUDA \textit{block} is a programming abstraction that represents a group of threads. Typically, a low threads/block setting will incur more natural barrier synchronizations that lead to higher latency, while too high threads/block invites higher communication overhead between threads within the block. Table~\ref{Table_NCU} shows GATSPI's kernels are no exception.

Lastly, GATSPI's kernel's theoretical maximum occupancy is 50\% because each thread uses more than 32 32-bit registers. Forcing the compiler to use no more than 32 registers/thread can boost the occupancy metric, but to detrimental effect on latency. In this case, the code will be compiled to more machine instructions (not shown in Table~\ref{Table_NCU}), L1 hit rate plummets as each thread has fewer registers to use, and data is frequently cycled between the regfile and L1 (i. e. register spilling).

\textit{\textbf{Scaling Analysis}}
~---We further conduct experiments where we cut out key functional features of GATSPI and re-measure the kernel runtimes (Table~\ref{Table_features}). Lines 11-12 in Algo. 1 are simply deleted to do away with inertial net delay filtering. The average rise/fall delays of each input-pin-to-output-pin arc across all conditional SDF arcs are implemented as 2 element arrays instead of the 2-D arrays in Fig.~\ref{Figs_WaveformLUTSDF} to implement GATSPI with partial SDF capabilities. Table~\ref{Table_features} assesses the overheads incurred from these key functional features, which is 5-13\% of total kernel runtime. Thus, it is a good tradeoff to include these features considering the accuracy requirements of industrial-grade power analysis.
\input{figs/Table_NSYS}
\input{figs/Table_NCU}
\input{figs/Table_features}
\input{figs/Table_archScaling}

Table~\ref{Table_archScaling} shows GATSPI performance across different NVIDIA GPUs, an A100 and a T4. Using a single GPU on an NVIDIA DGX Station A100 (with 40 GB memory), we find that A100 provides a 1.2-1.5X speedup over V100, scaling some benchmarks by more than the increase in SM count, likely due to larger increases in memory bandwidth and L2 cache size  (Table~\ref{Table_NCU} shows L2 hit rate is low). This is beneficial for industry-sized benchmarks, especially as designs become larger and testbenches more numerous, since these benchmarks are able to take advantage of memory capability improvements.
When scaling down to lower-cost GPUs (benchmarking on an NVIDIA T4 in an AWS g4dn.xlarge instance\cite{ProblemC}), we find that GATSPI runs 4.2-6.7X slower on a T4 than a V100.  The NVDLA(large),scan results show a 4.2X slowdown, exhibiting similar scaling to typical deep learning performance benchmarks\cite{T4Benchmarks}. The Design B experiments slow down more  because they require multiple sequential invocations of GATSPI---the full testbench waveforms did not fit in device memory.

Lastly, we benchmark performance of GATSPI in multi-GPU systems. We chose to implement a simple workload distribution strategy by distributing cycle parallelism across the GPUs. Namely, cycle parallelism is set to $32n$, where $n$ is number of GPUs, and each GPU is responsible for simulating 32 independent cycles in parallel. Although this form of workload distribution can be further enhanced with distribution across design parallelism with the incorporation of efficient netlist partitioning and/or duplication methods~\cite{Sen2010,ESSENT}, we leave that to future work. With our cycle parallelism workload distribution approach, the kernel runtime follows the equation $t=t_1/n+ovr$, where $t_1$ is single GPU runtime, and $ovr$ is the stream synchronize + kernel launch overhead. Fig.~\ref{Figs_multiGPU} shows the results of such an experiment where we concatenate all the testbenches in Table~\ref{Table_benchmarks} for Design B for re-simulation. We used a DGX Station V/A100 with 8/4 GPUs for the multi-GPU system runs. In single GPU systems, we must invoke GATSPI multiple times to complete the long concatenated testbench to ensure we don't surpass memory capacity. In multi-GPU systems, we distribute the workload across cycle parallelism as described above. Concatenating all testbenches helps amortize the $ovr$ cost. GATSPI achieves 7412X and 6672X kernel speedup on 8 V100s and 4 A100s, respectively. Deviation from perfect linear scaling is due to uneven activity factor between the distributed cycle-parallelism workloads.

\input{figs/Figs_multiGPU}

%% file: figs/Table_NSYS.tex
\begin{table}[t]
  \centering
  \captionsetup{belowskip=0pt,aboveskip=0pt}
  \caption{\normalfont{Nsight profiling of application runtime, in units of seconds.}}
    \resizebox{\columnwidth}{!}{\begin{tabular}{cccc}
    \textbf{Design(Testbench)} & \multicolumn{1}{M{5.255em}}{\textbf{Host to Device Data Transfer}} & \multicolumn{1}{M{6.58em}}{\textbf{Stream Synchronize + Kernel Launch}} & \multicolumn{1}{M{4.745em}}{\textbf{Kernel Execution}} \\
    \midrule
    Design A(func. 1) & 0.68  & 0.24  & 0.52 \\
    Design B(func. 2) & 3.40  & 3.50  & 10.80 \\
    Design B(high activity) & 7.82  & 3.50  & 31.34 \\
    \bottomrule
    \end{tabular}}
  \label{Table_NSYS}%
    \vspace{0pt}
\end{table}%

%% file: figs/Table_NCU.tex
\begin{table*}[t]
\centering
\captionsetup{belowskip=0pt,aboveskip=0pt}
\caption{\normalfont{Nsight profiling of re-simulation kernel on V100. Throughput \% numbers relative to maximum capability of GPU. Memory Throughput \% is a compound metric encasing L1, L2, and Global memory performance.}}
\label{Table_NCU}
\resizebox{\textwidth}{!}{
    \begin{tabular}{cc|ccccccc|cc}
    \textbf{Design(Testbench)} & \multicolumn{1}{M{6.265em}|}{\textbf{Config. \{Cycle parallelism, threads/block, regs/thread\}}} & \textbf{Threads} & \multicolumn{1}{M{5.49em}}{\textbf{Compute/ Memory Throughput (\%)}} & \textbf{Occupancy(\%)} & \multicolumn{1}{M{5.065em}}{\textbf{Global Memory Throughput (GB/s)}} & \multicolumn{1}{M{3.055em}}{\textbf{L1/L2 Hit Rate(\%)}} & \multicolumn{1}{M{4.555em}}{\textbf{Cycles per Scheduler Issue}} & \multicolumn{1}{M{5.19em}|}{\textbf{Uncoalesced Memory Accesses(\%)}} & \multicolumn{1}{M{3.055em}}{\textbf{Elapsed GPU Cycles}} & \multicolumn{1}{M{3.055em}}{\textbf{Latency (ms)}} \\
    \midrule
    \textcolor[rgb]{ 0,  0,  1}{\textbf{Design A(func. 1)}} & \textcolor[rgb]{ 0,  0,  1}{32,512,64} & \textcolor[rgb]{ 0,  0,  1}{170k} & \textcolor[rgb]{ 0,  0,  1}{10.4/12.1} & \textcolor[rgb]{ 0,  0,  1}{28.8} & \textcolor[rgb]{ 0,  0,  1}{7.0} & \textcolor[rgb]{ 0,  0,  1}{92.4/94.4} & \textcolor[rgb]{ 0,  0,  1}{2.4} & \textcolor[rgb]{ 0,  0,  1}{48} & \textcolor[rgb]{ 0,  0,  1}{6.6M} & \textcolor[rgb]{ 0,  0,  1}{6.00} \\
    \textbf{Design A(func. 1)} & 128,512,64 & 680k  & 32.7/32.9 & 38.7  & 28.5 & 94.8/87.6 & 2.0   & 18    & 3.3M  & 3.28 \\
    \textbf{Design A(func. 1)} & 256,512,64 & 1.36M & 44.5/46.1 & 41.8  & 39.5  & 96.2/82.0 & 2.0   & 12    & 3.8M  & 3.29 \\
    \midrule
    \textcolor[rgb]{ 0,  0,  1}{\textbf{Design B(func. 2)}} & \textcolor[rgb]{ 0,  0,  1}{32,512,64} & \textcolor[rgb]{ 0,  0,  1}{4.1M} & \textcolor[rgb]{ 0,  0,  1}{33.8/44.6} & \textcolor[rgb]{ 0,  0,  1}{42.9} & \textcolor[rgb]{ 0,  0,  1}{144.2} & \textcolor[rgb]{ 0,  0,  1}{91.5/60.0} & \textcolor[rgb]{ 0,  0,  1}{2.8} & \textcolor[rgb]{ 0,  0,  1}{23} & \textcolor[rgb]{ 0,  0,  1}{380.7M} & \textcolor[rgb]{ 0,  0,  1}{335.76} \\
    \textcolor[rgb]{ 0,  0,  1}{\textbf{Design B(high activity)}} & \textcolor[rgb]{ 0,  0,  1}{32,512,64} & \textcolor[rgb]{ 0,  0,  1}{4.1M} & \textcolor[rgb]{ 0,  0,  1}{28.4/38.4} & \textcolor[rgb]{ 0,  0,  1}{48.0} & \textcolor[rgb]{ 0,  0,  1}{281.1} & \textcolor[rgb]{ 0,  0,  1}{84.5/47.0} & \textcolor[rgb]{ 0,  0,  1}{3.5} & \textcolor[rgb]{ 0,  0,  1}{41} & \textcolor[rgb]{ 0,  0,  1}{788.3M} & \textcolor[rgb]{ 0,  0,  1}{696.36} \\
    \textbf{Design B(high activity)} & 64,512,64 & 8.2M  & 33.1/40.8 & 48.3  & 241.4 & 86.8/52.4 & 3.0   & 23    & 871.6M & 772.96 \\
    \textbf{Design B(high activity)} & 128,512,64 & 16.3M & 39.5/47.1 & 48.8  & 158.1 & 90.9/55.7 & 2.6   & 12    & 1.1B  & 993.69 \\
    \textbf{Design B(high activity)} & 32,1024,64 & 4.1M  & 27.6/37.1 & 49.8  & 270.8 & 87.1/49.9 & 3.5   & 52    & 878.8M & 776.85 \\
    \textbf{Design B(high activity)} & 32,512,32 & 4.1M  & 15.4/46.1 & 94.4  & 402.8 & 68.1/52.2 & 6.4   & 42    & 1.5B  & 1350.00 \\
    \bottomrule
    \end{tabular}%
}
\vspace{-15pt}
\end{table*}

%% file: figs/Table_features.tex
\begin{table}[t]
  \centering
  \captionsetup{belowskip=0pt,aboveskip=0pt}
  \caption{\normalfont{Kernel runtimes (s) and speedups without key features.}}
    \resizebox{\columnwidth}{!}{\begin{tabular}{cccc}
    \textbf{Design(Testbench)} & \multicolumn{1}{M{4.99em}}{\textbf{Full Features}} & \multicolumn{1}{M{4.98em}}{\textbf{No Net Delay}} & \multicolumn{1}{M{4.935em}}{\textbf{No Net Delay + No Full SDF}} \\
    \midrule
    Design A(func. 1) & 0.79(808X) & 0.79(808X) & 0.70(907X) \\
    Design B(func. 2) & 14.55(1026X) & 14.19(1052X) & 13.80(1081X) \\
    Design B(high activity) & 38.90(1198X) & 36.03(1294X) & 33.65(1385X) \\
    \bottomrule
    \end{tabular}}
  \label{Table_features}
    \vspace{2pt}
\end{table}%

%% file: figs/Table_archScaling.tex
\begin{table}[t]
  \centering
  \captionsetup{belowskip=0pt,aboveskip=3pt}
  \caption{\normalfont{Kernel runtimes (s) and speedups for different GPUs.}}
    \resizebox{\columnwidth}{!}{\begin{tabular}{cccc}
    \textbf{Design(Testbench)} & \multicolumn{1}{M{4.555em}}{\textbf{T4}} & \multicolumn{1}{M{3.89em}}{\textbf{V100}} & \multicolumn{1}{M{4.22em}}{\textbf{A100}} \\
    \midrule
    NVDLA,large(scan) & 42.53(60X) & 9.99(254X) & 6.59(385X) \\
    Design B(func. 2) & 76.56(195X) & 14.55(1026X) & 12.11(1232X) \\
    Design B(high activity) & 260.15(179X) & 38.90(1198X) & 25.50(1828X) \\
    \bottomrule
    \end{tabular}}
  \label{Table_archScaling}%
    \vspace{2pt}
\end{table}%

%% file: figs/Figs_multiGPU.tex
\begin{figure}[t]
  \setlength{\abovecaptionskip}{0pt}
  \setlength{\belowcaptionskip}{0pt}
  \centering
  \includegraphics[width=1\columnwidth]{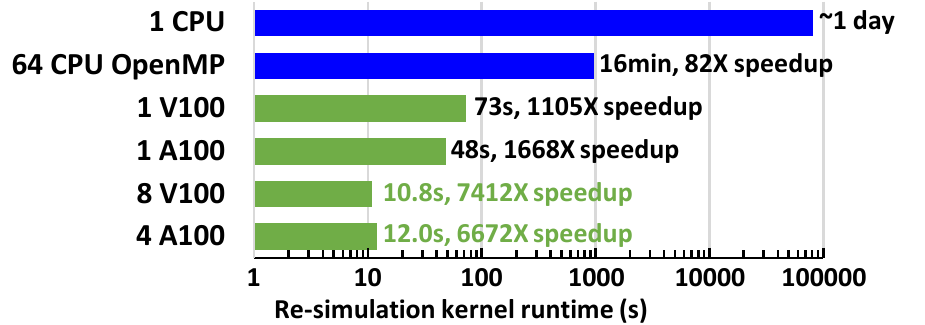}
  \caption{\normalfont{Re-simulation kernel runtime comparisons across different hardware platforms for Design B.}}
  \label{Figs_multiGPU}
\end{figure}

%% file: 6.Conclusions.tex
\section{Conclusions and Future Work}
\label{6.Conclusions}
In this paper, we presented GATSPI, a novel PyTorch-written GPU accelerated logic gate re-simulator that enables ultra-fast power estimation (28-1198X kernel speedup, 5-680X application speedup) across a diverse set of benchmarks. Re-simulation results match a commercial simulator baseline for accuracy, an important criteria in industry power analysis use cases. GATSPI incorporates important re-simulator functional features such as support for a wide range of cell types, SDF statements, and inertial delay filtering. With no calibration, GATSPI exhibits robust runtimes across benchmarks. These advantages motivate GATSPI to be deployed in real-world applications such as glitch-optimization loops, where we demonstrate 1.4\% power savings with 449X speedup.

Our analysis provides insight and opportunities for future work. Inclusion of sequential element simulation can be explored to see if full logic simulation could provide similar acceleration potential. Memory management techniques can be incorporated to alleviate memory pressure for further scaling. Optimizations to exploit design parallelism, such as more sophisticated netlist partitioning techniques may help benchmarks with unbalanced activity workloads~\cite{ESSENT} and shift GPU accelerated simulation more towards the event-based paradigm. Finally, GATSPI's ultra-fast speeds open the door for exploration of its usefulness in additional applications, such as crosstalk analysis, IR drop analysis, and fault simulation.  GATSPI could also be a valuable tool in the development of GPU-accelerated AI-assisted EDA algorithms.

%% file: main.bbl
\begin{thebibliography}{10}

\bibitem{ProblemC}
Y.~Zhang, H.~Ren, B.~Keller, and B.~Khailany.
\newblock {Problem C: GPU Accelerated Logic Re-simulation : (Invited Talk)}.
\newblock In {\em ICCAD}, 2020.

\bibitem{Holst2015}
S.~Holst, M.~E. Imhof, and H.-J. Wunderlich.
\newblock {High-Throughput Logic Timing Simulation on GPGPUs}.
\newblock In {\em TODEAS}, June 2015.

\bibitem{Zhu2011}
Y.~Zhu, B.~Wang, and Y.~Deng.
\newblock {Massively Parallel Logic Simulation with GPUs}.
\newblock In {\em TODAES}, June 2011.

\bibitem{Sen2010}
A.~Sen, B.~Aksanli, M.~Bozkurt, and M.~Mert.
\newblock {Parallel Cycle Based Logic Simulation Using Graphics Processing
  Units}.
\newblock In {\em ISPDC}, 2010.

\bibitem{Chatterjee2011}
D.~Chatterjee, A.~Deorio, and V.~Bertacco.
\newblock {Gate-Level Simulation with GPU Computing}.
\newblock In {\em TODAES}, 2011.

\bibitem{Lai2020}
L.~Lai, Q.~Zhang, H.~Tsai, and W.-T. Cheng.
\newblock {GPU-based Hybrid Parallel Logic Simulation for Scan Patterns}.
\newblock In {\em ITC-Asia}, 2020.

\bibitem{Zeng2021}
C.~Zeng, F.~Yang, and X.~Zeng.
\newblock {Accelerate Logic Re-simulation on GPU via Gate/Event Parallelism and
  State Compression}.
\newblock In {\em ICCAD}, 2021.

\bibitem{GPUarchs}
{NVIDIA A100 Tensor Core GPU Architecture}.
\newblock {\em
  https://images.nvidia.com/aem-dam/en-zz/Solutions/data-center/nvidia-ampere-architecture-whitepaper.pdf},
  2020.

\bibitem{T4arch}
{T4 Tensor Core Product Brief}.
\newblock {\em
  https://www.nvidia.com/content/dam/en-zz/Solutions/Data-Center/tesla-t4/t4-tensor-core-product-brief.pdf},
  2020.

\bibitem{PyTorch}
A.~Paszke~et al.
\newblock {PyTorch: An Imperative Style, High-Performance Deep Learning
  Library}.
\newblock In {\em NeurIPS}. 2019.

\bibitem{DGL}
M.~Wang~et al.
\newblock {Deep Graph Library: A Graph-Centric, Highly-Performant Package for
  Graph Neural Networks}.
\newblock {\em arXiv preprint arXiv:1909.01315}.

\bibitem{Lin2019}
Y.~Lin~et al.
\newblock {DREAMPlace: Deep Learning Toolkit-Enabled GPU Acceleration for
  Modern VLSI Placement}.
\newblock In {\em DAC}, 2019.

\bibitem{GRANNITE}
Y.~Zhang, H.~Ren, and B.~Khailany.
\newblock {GRANNITE: Graph Neural Network Inference for Transferable Power
  Estimation}.
\newblock In {\em DAC}, 2020.

\bibitem{ICCAD2020Winner}
S.~M. Sadati, M.~Shahidzade, and B.~Ghavami.
\newblock {ICCAD 2020 CAD Contest Video Shows of Problem C: cada0143}.
\newblock {\em http://iccad-contest.org/2020/winners.html}, 2020.

\bibitem{CUDAtuning}
G.~Thomas-Collignon and V.~Mehta.
\newblock {Optimizing Applications for NVIDIA Ampere GPU Architecture}.
\newblock In {\em GTC}, 2020.

\bibitem{NVDLA}
{NVDLA Open Source Project}.
\newblock {\em http://nvdla.org/}, 2019.

\bibitem{nsys}
{NVIDIA Nsight}.
\newblock {\em https://developer.nvidia.com/nsight-visual-studio-edition}.

\bibitem{T4Benchmarks}
{Deep Learning Performance on T4 GPUs with MLPerf Benchmarks}.
\newblock {\em
  https://www.dell.com/support/kbdoc/en-us/000132094/deep-learning-performance-on-t4-gpus-with-mlperf-benchmarks},
  2021.

\bibitem{ESSENT}
S.~Beamer and D.~Donofrio.
\newblock {Efficiently Exploiting Low Activity Factors to Accelerate RTL
  Simulation}.
\newblock In {\em DAC}, 2020.

\end{thebibliography}
